\documentclass[11pt,a4paper]{article}

\usepackage[final]{acl}

\usepackage{times}
\usepackage{latexsym}
\usepackage{hyperref}
\usepackage{multirow}
\usepackage{float}
\usepackage{caption}
\usepackage{tabularx}
\usepackage[table]{xcolor}

\usepackage[T1]{fontenc}

\usepackage[utf8]{inputenc}

\usepackage{microtype}

\usepackage{inconsolata}

\usepackage{graphicx}
\usepackage{booktabs}
\graphicspath{ {./figures/} }

\title{Validate Your Authority: Benchmarking LLMs on Multi-Label Precedent Treatment Classification}

\author{\textbf{M. Mikail Demir} and \textbf{M. Abdullah Canbaz} \\
        Department of Information Science and Technology \\ College of Emergency Preparedness, Homeland Security, and Cybersecurity \\ University at Albany, SUNY \\ mdemir, mcanbaz [at] albany [dot] edu}

\begin{document}
\maketitle
\begin{abstract}
Automating the classification of negative treatment in legal precedent is a critical yet nuanced NLP task where misclassification carries significant risk. To address the shortcomings of standard accuracy, this paper introduces a more robust evaluation framework. We benchmark modern Large Language Models on a new, expert-annotated, publicly available dataset of 239 real-world legal citations and propose a novel Average Severity Error metric to better measure the practical impact of classification errors. Our experiments reveal a performance split: Google's Gemini 2.5 Flash achieved the highest accuracy on a high-level classification task (79.1\%), while OpenAI's GPT-5-mini was the top performer on the more complex fine-grained schema (67.7\%). This work establishes a crucial baseline, provides a new context-rich dataset, and introduces an evaluation metric tailored to the demands of this complex legal reasoning task.
\end{abstract}

\section{Introduction}

In common law jurisdictions, the doctrine of precedent, or \textit{stare decisis}, is a cornerstone of the legal system \citep{AmericanBarAssociation_Undated_UnderstandingStare}. It compels judges to decide cases by referencing previous decisions with similar factual situations. This makes it imperative for legal practitioners to determine if a judicial decision is still considered "good law," or if its authority has been weakened or nullified by subsequent cases. A case can be "negatively treated" in various ways; for instance, it can be explicitly "overruled" by a higher court, its reasoning can be "criticized," or it can be "distinguished" as not applying to a new set of facts. For a lawyer, building an argument upon a case that is no longer good law can be a critical error, potentially leading to unfavorable judgments and professional repercussions—especially when the error is severe.

To address this challenge, the legal industry has historically relied on commercial citator services. Leading platforms like Shepard's on LexisNexis \citep{lexis_shepards_29step_2014,lexis_shepards_history_2022}, KeyCite on Westlaw \citep{westlaw_editorial_enhancements_2025,westlaw_attorney_editors_video_2025}, and BCite on Bloomberg Law \citep{bloomberg_bcite_help_center_2025,bloomberg_bcite_litigators_pdf_2021} employ teams of legal editors to analyze how cases are treated in subsequent decisions. These services provide signals, such as color-coded flags, to quickly alert practitioners to potential negative treatment\footnote{An example of use of color codes and descriptive labels by Shepard's, citation product of LexisNexis, a division of RELX Inc., can be found here \url{https://supportcenter.lexisnexis.com/app/answers/answer_view/a_id/1088155/~/shepards-signals-and-analysis}}. However, these services are not infallible.

Documented human error in manual citation analysis  \cite{hellyerEvaluatingShepardsKeyCite2018a} have spurred interest in automated solutions. The recent emergence of powerful Large Language Models (LLMs) has significantly advanced the capabilities of legal natural language processing (NLP), offering the potential to handle the complex, nuanced reasoning inherent in legal texts.

This paper builds upon this body of work by evaluating the performance of modern LLMs on the task of classifying case law citation treatments. The central research question we address is:

\textbf{How accurately can contemporary LLMs replicate the sophisticated legal reasoning required to classify the various ways a judicial precedent is treated in a subsequent case?} 

Early research in this area utilized rule-based approaches \citep{sartorNormativeConflictsLegal1992,prakkenLogicalFrameworkModelling1993}. Later work, such as the LEXA system by Galgani and Hoffmann (2010), demonstrated the feasibility of combining knowledge engineering with baseline machine learning models for this task. More recently, traditional machine learning models \cite{lockeAutomaticallyClassifyingCase2019a} were used to classify how a cited case is treated. 

While previous studies have highlighted the intrinsic difficulty of this task for earlier neural network architectures, the capabilities of the latest generation of LLMs remain to be thoroughly benchmarked. This paper contributes to the field by presenting a systematic evaluation of these models on a challenging, multi-label classification task, using an expert-annotated dataset of real-world legal decisions to provide a new baseline for this critical task.

\section{Related Work}
In this section, we are reporting the related work to the research question that we are examining in this paper, in 3 category that surrounds the task that we have defined with our research question. 
\subsection{Comparative Studies of Commercial Citators}
The task of validating case law is critical for legal practice, yet manual review is fraught with challenges. Seminal studies by \citep{taylorComparingKeyCiteShepards2000} and \citep{hellyerEvaluatingShepardsKeyCite2018a} put the major commercial citator services to the test. Their findings revealed significant rates of error and inconsistency, with the services missing or mislabeling a substantial portion of negative citation treatments.
\subsection{Machine Learning Enabled Legal Reasoning and Classification}
The legal field has long been a target for automation through NLP. Domain-specific models such as LEGAL-BERT \citep{chalkidisLEGALBERTMuppetsStraight2020},when pre-trained on large legal corpora, showed improved performance on complex legal tasks. One example is by \citet{zhengWhenDoesPretraining2021}, which tried to identify case holdings from the CaseHOLD dataset which resembles with the work at hand.

However, the specific task of classifying citation treatment has remained a significant challenge. Early work by \citet{galganiLEXAAutomaticLegal2010} demonstrated the task's complexity, a finding reinforced by a notable study from \citet{lockeAutomaticallyClassifyingCase2019a} which investigated various neural network architectures for this purpose and found the task to be intrinsically difficult, with most models performing poorly. This established an important benchmark for the complexity of the task before the widespread availability of modern LLMs.

The recent advent of Transformer-powered LLMs is now fundamentally transforming the legal sector. These models are pioneering change by automating intricate tasks such as predicting legal judgments \citep*{chalkidisNeuralLegalJudgment2019}, analyzing vast legal documents \citep{mamakasProcessingLongLegal2022}. This development also holds hope for democratizing legal services and addressing the global access-to-justice crisis \citep{GenerativeAILegal2021}.

Despite this potential, applying LLMs in law is constrained by major factors. The LegalBench benchmark put by \citet{guhaLegalBenchCollaborativelyBuilt2023a}, reports the broad capabilities of LLMs, also highlights the challenge of data scarcity, as high-quality, expert-annotated legal data is expensive and difficult to produce . Furthermore, the use of external, third-party LLMs raises critical privacy and confidentiality concerns \citep{demirLegalGuardianPrivacyPreservingFramework2025}.

To our knowledge, this paper is the first academic study to systematically benchmark modern LLMs on the fine-grained, multi-label classification of negative citation treatments. We aim to establish a new performance baseline for this fundamental legal reasoning task, addressing a known hard problem within the current LLM landscape.

\section{Methodology}
\subsection{Dataset Curation}
A primary challenge in developing and evaluating specialized legal NLP applications is the scarcity of high-quality, annotated data. While general legal benchmarks like LegalBENCH exist \citep{guhaLegalBenchCollaborativelyBuilt2023a}, and specific datasets like CaseHOLD \cite{zhengWhenDoesPretraining2021} address tasks such as identifying overruled cases, they often have limitations for our specific purpose. CaseHOLD, for instance, provides a binary classification (overruled or not) based on a single holding sentence, lacking the broader case context and the fine-grained, multi-label classifications needed to rigorously test the nuanced reasoning capabilities of LLMs.

To address this gap, we build on the annotated corpus from \citet{hellyerEvaluatingShepardsKeyCite2018a}, which empirically evaluated commercial citators. The author shared the annotations for research use, providing an ideal foundation for our experiments.

To utilize this expert analysis for our computational experiments, we undertook a multi-stage process to structure, enrich, and filter the data. The initial corpus, provided in PDF format, was first digitized manually into a structured CSV file. Following \citet{hellyerEvaluatingShepardsKeyCite2018a}'s methodology, we excluded citing relationships that were marked as ambiguous (cases where reasonable legal experts might disagree on the treatment) resulting in a high-confidence set of 329 citing relationships. An example of a entry of a dataset in the format provided by Hellyer is provided in the Appendix \ref{sec:B}, Figure \ref{fig:dataset_example}.

To create a definitive ground truth from the remaining annotations, we developed a systematic, priority-based logic. The highest priority was assigned to entries \citeauthor{hellyerEvaluatingShepardsKeyCite2018a} marked as an explicit correction (e.g., a commercial citator's "\textit{Criticized by}" label corrected to "[\textit{Not Followed}]"). In these instances, the corrected label within the brackets was adopted as the single, most accurate ground truth. An example of an explicit correction from the dataset is provided in the Appendix \ref{sec:B}, Figure \ref{fig:dataset_example_bracket}.

\begin{figure*}
    \centering
    \includegraphics[width=\textwidth]{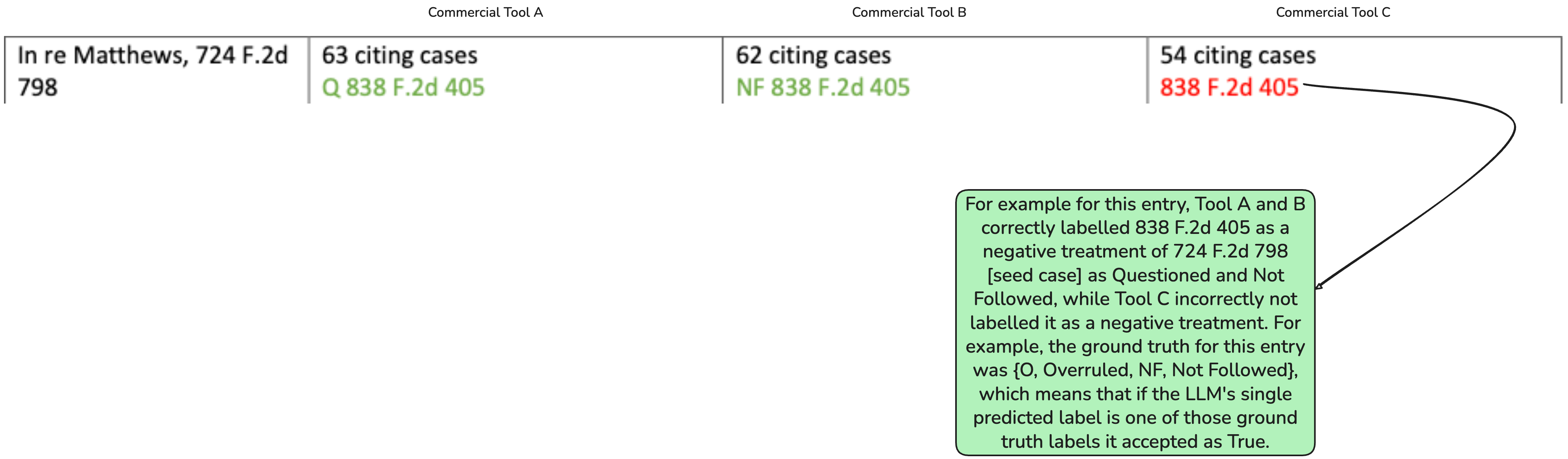}     
    \caption{A snippet of the dataset that \cite{hellyerEvaluatingShepardsKeyCite2018a} provided, where more than one label is accepted as ground truth
}
    \label{fig:dataset_example_multi}
\end{figure*}

For entries without an explicit correction, we interpreted labels marked as "acceptable" as our ground truth. If multiple, different labels were deemed acceptable for a single relationship (e.g., both "\textit{Criticized by}" and "\textit{Questioned} by"), our ground truth embraced this nuance by including both labels. We then considered the LLM's single-label prediction to be correct if it matched any of the acceptable ground truth labels. An example of a relationship with multiple acceptable labels has also been provided in the \ref{sec:B}, Figure \ref{fig:dataset_example_multi}.

Finally, we retrieved citing-case full text via the CourtListener REST API\footnote{\url{CourtListener REST API: https://www.courtlistener.com/api/rest/v4/}} to provide models with full-document context. CourtListener is the public-facing access point for the Free Law Project, a non-profit initiative dedicated to providing free, public access to primary legal materials. As a public-benefit, non-commercial repository, its collection, while vast, is not exhaustive and can have gaps compared to proprietary legal databases. We encountered two primary challenges in our data retrieval process: some cases in our set were not yet included in CourtListener's public collection, and other citations were provided as LexisNexis citation slips, which are not universally indexed and are therefore difficult for public initiatives like the Free Law Project to resolve.

We programmatically queried the CourtListener API for each of the 329 citing relationships in our structured dataset. As of August 13, 2025, we successfully retrieved full-text opinions for 239 citing cases. This final corpus of 239 fully-contextualized citing relationships forms the basis for our evaluation. In the subsequent section, we detail the key statistical properties of this dataset, including the distribution of ground truth labels, which reveals a significant class imbalance inherent to real-world legal data; the textual complexity of the citing documents, measured by their average token count; and the multi-label nature of the annotations, a direct result of \citet{hellyerEvaluatingShepardsKeyCite2018a}'s methodology allowing for multiple acceptable interpretations of a single legal treatment.

\subsection{Label Distribution and Classification Schema}

The final dataset of 239 citing relationships forms the basis for our evaluation. Each data point was structured to provide the full context necessary for complex legal reasoning. An example of a single, fully-processed entry is shown in Table \ref{tab:dataset-variables}.

\begin{table}[h!]
    \centering
    \begin{tabular}{@{}l@{}}
        \toprule
        \textbf{Variable Name} \\ 
        \midrule
        Seed Case Citation \\
        Seed Case Name \\
        Citing Case Citation \\
        Citing Case Text \\
        True Label (FG) \\
        True Label (HL) \\ 
        \bottomrule
    \end{tabular}
    \caption{Variables for each citing relationship.}
    \label{tab:dataset-variables}

\end{table}

A summary of these properties is presented in Table \ref{tab:accents},\ref{tab:labels_distribution} and \ref{tab:high_level_labels_distribution}. The citing documents are textually complex, with an average length of over 7,000 tokens, requiring models to process substantial context to identify the relevant legal treatment.
\begin{table}[h]
  \centering
  \begin{tabular}{lc}
    \hline
    \textbf{Summary Statistic} & \textbf{Count} \\
    \hline
    Total Citing Relationship     & 239 \\
    Average Token Count     & 7296 \\ 
    Relationships with \textit{>1} True Label     & 46 \\ 
    
    \hline
  \end{tabular}
  \caption{Summary statistics for the corpus}
  \label{tab:accents}
\end{table}

A defining feature of this corpus, inherited from \citeauthor{hellyerEvaluatingShepardsKeyCite2018a}'s methodology, is its multi-label nature. The annotation process was not designed to find a single, objectively "correct" label, but rather to identify all treatments that a reasonable legal expert would find acceptable. As \citeauthor{hellyerEvaluatingShepardsKeyCite2018a} notes, it is common for different citators to apply different descriptive labels to the same citing relationship. Our ground truth embraces this nuance by allowing multiple valid labels for a single case, a characteristic reflected in our data where more than 19\% of relationships have more than one fine-grained label.

The distribution of these labels reveals a significant class imbalance inherent to the dataset that we have utilized. The dataset is skewed towards negative treatments because it was constructed exclusively from cases already flagged as negative. As a result, less severe labels such as LIMITED OR DISTINGUISHED\footnote{Within the rest of this work, fine-grained labels typed as \textit{\textbf{This}}, while high-level labels typed as \textbf{THIS}.} (and its fine-grained counterpart, \textit{Distinguished}) are far more common than severe, dispositive treatments like INVALIDATED. 

Notably, the presence of the non-negative labels is a direct artifact of the dataset's origin. These instances represent cases that commercial citators incorrectly flagged as negative, which Hellyer subsequently corrected to non-negative treatments. We have included this entries as neutral citations to provide a testbed for evaluating an LLM's ability to reject these false positives.

\begin{table}[ht]
  \centering
  
  \begin{tabular}{lc}
    \hline
    \textbf{Label} & \textbf{Count} \\
    \hline
    \textit{Distinguished by} & 132 \\
    \textit{Criticized by} & 36 \\
    \textit{Not followed by} & 21 \\
    \textit{Overruling recognized by} & 18 \\
    \textit{Neutral} & 10 \\
    \textit{Disagreement recognized by} & 10 \\
    \textit{Disagreed with by} & 10 \\
    \textit{Questioned by} & 9 \\
    \textit{Declined to extend by} & 9 \\
    \textit{Among conflicting authorities noted in} & 8 \\
    \textit{Called into doubt by} & 8 \\
    \textit{Overruled} & 6 \\
    \textit{Abrogation recognized by} & 4 \\
    \textit{Reversed by} & 2 \\
    \textit{Implied overruling recognized by} & 1 \\
    \textit{Disapproved as stated in} & 1 \\
    \textit{Limitation of holding recognized by} & 1 \\
    \hline
    \textbf{Total} & \textbf{286} \\
    \hline
  \end{tabular}
  \caption{Distribution of fine-grained ground-truth labels (multi-label; totals exceed 239 relationships)}
  \label{tab:labels_distribution}
  
  \vspace{1cm} 

  \begin{tabular}{lc}
    \hline
    \textbf{Label} & \textbf{Count} \\
    \hline
    LIMITED OR DISTINGUISHED & 156 \\
    CRITICIZED OR QUESTIONED & 49 \\
    INVALIDATED & 30 \\
    CONFLICT NOTED & 16 \\
    NEUTRAL CITATION & 10 \\
    \hline
    \textbf{Total} & \textbf{261} \\
    \hline
  \end{tabular}
  \caption{Distribution of high-level ground-truth labels (multi-label; totals exceed 239 relationships)}
  \label{tab:high_level_labels_distribution}

\end{table}

While the fine-grained schema offers high precision, its 16 distinct categories can pose a challenge for both model classification and high-level analysis. To facilitate a broader understanding of precedent treatment, we developed a hierarchical, high-level classification schema that groups semantically similar fine-grained labels. This schema, detailed in Table \ref{tab:schema_mapping_detailed}, condenses the fine-grained categories into five conceptually distinct groups, allowing for a clearer interpretation of the models' core reasoning capabilities. For instance, all labels indicating direct negative commentary on a case's reasoning are grouped under CRITICIZED OR QUESTIONED, while all labels that limit a precedent's scope without nullifying it are grouped under LIMITED OR DISTINGUISHED. This two-tiered schema allows for a comprehensive evaluation at both a granular and a conceptual level.

\begin{table*}[ht]
\centering
\begin{tabular}{ll}
\toprule
\textbf{High-Level Category} & \textbf{Fine-Grained Label} \\
\midrule
\textbf{INVALIDATED} & Overruled \\
&Overruling recognized by \\
&Implied overruling recognized by \\
&Abrogation recognized by \\
&Reversed by \\
\midrule
\textbf{CRITICIZED OR QUESTIONED} & Criticized by \\
& Called into doubt by \\
& Questioned by \\
& Disagreed with by \\
& Disapproved as stated in \\
\midrule
\textbf{LIMITED OR DISTINGUISHED}& Distinguished by \\
& Declined to extend by \\
& Limitation of holding recognized by \\
& Not followed by \\
\midrule
\textbf{CONFLICT NOTED} & Among conflicting authorities noted in \\
& Disagreement recognized by \\
\midrule
\textbf{NEUTRAL CITATION}& Neutral Citation \\
\bottomrule
\end{tabular}
\caption{The hierarchical mapping from fine-grained to high-level labels. This table defines each fine-grained abbreviation and its mapping to the five high-level categories.}
\label{tab:schema_mapping_detailed}
\end{table*}
\section{Experiment}
To evaluate the capabilities of modern LLMs on this task, we designed a systematic experimental setup focusing on zero-shot and few-shot learning paradigms. This approach was chosen to simulate a realistic use case where fine-tuning a model on a large, domain-specific dataset is often impractical due to data scarcity and computational cost.

Our experiments include a representative sample of both proprietary and high-performance open-source models. For proprietary models, we selected Google's Gemini 2.5 Pro and Gemini 2.5 Flash, alongside OpenAI's GPT-5-mini, all accessed via their official APIs. Our choice was guided by the availability of research credits, a common constraint in academic research; a detailed discussion of this limitation is provided in the \hyperlink{sec:limitations}{Limitations and Future Work} section. For the open-source model, we evaluated the Qwen3 (30B variant), which was served using vLLM on a machine equipped with two NVIDIA RTX A6000 GPUs, allowing us to evaluate its full-precision bfloat16 version.

\subsection*{Prompting Strategies}
Our experimental design centers on two primary prompting strategies: zero-shot and few-shot prompting. For each LLM call, the model was provided with a prompt constructed from each variable in citing relationship, listed in Table \ref{tab:dataset-variables}. Due to page constraints, detailed content of prompts used for our zero-shot and few-shot experiments are provided in Appendix \ref{sec:prompts}. 

We constructed our few-shot prompts by randomly selecting three precedent treatment examples from our dataset. To ensure a rigorous evaluation and prevent data contamination, any example used for in-context learning was excluded from the test set. For all prompts, the context provided to the model was a curated snippet from the citing case, specifically the paragraph(s) where the cited case is analyzed. We did not use the full text of the legal document. This was done for brevity and to isolate the most relevant text for the classification task.

\section{Results}
\subsection{Main Results}
In Table \ref{tab:main_results}, we report overall performance for each model under both the high-level and fine-grained schemas. The table includes the license type for each model to distinguish between proprietary and open-source systems. As the results indicate, proprietary models generally outperform the open-source Qwen3 model in this task. The best performance for the high-level schema was achieved by Gemini 2.5 Flash (Accuracy: 0.7908), while GPT-5-mini performed best on the more complex fine-grained schema (Accuracy: 0.6771), with top scores for each highlighted in bold.
\begin{table*}[t]
\centering
\begin{tabular}{lllcc}
\toprule
& & & \multicolumn{1}{c}{\textbf{High-Level }} & \multicolumn{1}{c}{\textbf{Fine-Grained }} \\
\cmidrule(lr){4-4} \cmidrule(lr){5-5}
\textbf{Model} & \textbf{License Type} & \textbf{Prompt Type} & \textbf{Accuracy} & \textbf{Accuracy} \\
\midrule
\multirow{2}{*}{Gemini 2.5 Flash} & \multirow{2}{*}{Proprietary} & Zero-Shot & \textbf{0.7908} & 0.6463 \\
                 &                              & Few-Shot  & 0.7699 & 0.6276 \\
\midrule
\multirow{2}{*}{Gemini 2.5 Pro}   & \multirow{2}{*}{Proprietary} & Zero-Shot & 0.7029 & 0.6638 \\
                 &                              & Few-Shot  & 0.7327 & 0.6682 \\
\midrule
\multirow{2}{*}{Qwen3:30B}        & \multirow{2}{*}{Apache 2.0}  & Zero-Shot & 0.6946 & 0.5356 \\
                 &                              & Few-Shot  & 0.5346 & 0.5484 \\
\midrule
\multirow{2}{*}{GPT-5-mini}        & \multirow{2}{*}{Proprietary} & Zero-Shot & 0.7597 & \textbf{0.6771} \\
                 &                              & Few-Shot  & 0.7005 & 0.5760 \\
\bottomrule
\end{tabular}
\caption{Overall performance of all models and prompt types on the high-level and fine-grained classification tasks. The primary metric shown is Instance-Based Accuracy. The best performing result in each schema is highlighted in bold.}
\label{tab:main_results}
\end{table*}
\subsection{Per-Label Performance Analysis}
To provide a more granular view of the top-performing models' capabilities, we present a detailed breakdown of their per-label classification performance for the best-performing model on each schema in Table \ref{tab:hl_details} and Table \ref{tab:fg_details_full_name}.

\begin{table*}[t]
\centering

\begin{tabular}{lcccc}
\toprule
\textbf{Label} & \textbf{Precision} & \textbf{Recall} & \textbf{F1-Score} & \textbf{Num. of Samples(Support)} \\
\midrule
LIMITED OR DISTINGUISHED & 0.947 & 0.795 & 0.864 & 156 \\
CRITICIZED OR QUESTIONED & 0.794 & 0.551 & 0.651 & 49 \\
INVALIDATED                & 0.840 & 0.700 & 0.764 & 30 \\
CONFLICT NOTED              & 0.353 & 0.750 & 0.480 & 16 \\
NEUTRAL CITATION          & 0.333 & 0.500 & 0.400 & 10 \\
\midrule
\textbf{Weighted Avg}      & \textbf{0.846} & \textbf{0.724} & \textbf{0.771} & \textbf{261} \\ 
\bottomrule
\end{tabular}
\caption{Per-label performance metrics for the best-performing model on the high-level schema (Gemini 2.5 Flash, Zero-Shot). Support refers to the number of true instances for each label.}
\label{tab:hl_details}
\end{table*}

\begin{table*}[t]
\centering
\small 
\begin{tabular}{lcccc}
\toprule
\textbf{Fine-Grained Label} & \textbf{Precision} & \textbf{Recall} & \textbf{F1-Score} & \textbf{Num. of Samples(Support)} \\
\midrule
Distinguished by (D) & 0.954 & 0.825 & 0.885 & 126 \\
Overruling recognized by (OR) & 1.000 & 0.722 & 0.839 & 18 \\
Disagreed with by (DW) & 0.471 & 0.800 & 0.593 & 10 \\
Overruled (O) & 0.667 & 0.333 & 0.444 & 6 \\
Among conflicting authorities noted in (ACAN) & 0.292 & 0.875 & 0.438 & 8 \\
Declined to extend by (DE) & 1.000 & 0.250 & 0.400 & 8 \\
Called into doubt by (CID) & 1.000 & 0.250 & 0.400 & 8 \\
Not followed by (NF) & 0.667 & 0.286 & 0.400 & 21 \\
Implied overruling recognized by (IOR) & 0.200 & 1.000 & 0.333 & 1 \\
Disagreement recognized by (DR) & 0.333 & 0.200 & 0.250 & 10 \\
Criticized by (C) & 0.364 & 0.111 & 0.170 & 36 \\
Reversed by (R) & 0.000 & 0.000 & 0.000 & 2 \\
Questioned by (Q) & 0.000 & 0.000 & 0.000 & 9 \\
Neutral Citation (N) & 0.000 & 0.000 & 0.000 & 0 \\
Limitation of holding recognized by (LHR) & 0.000 & 0.000 & 0.000 & 1 \\
Disapproved as stated in (DAS) & 0.000 & 0.000 & 0.000 & 1 \\
Abrogation recognized by (AR) & 0.000 & 0.000 & 0.000 & 4 \\
\midrule
\textbf{Weighted Avg} & \textbf{0.728} & \textbf{0.561} & \textbf{0.604} & \textbf{269} \\
\bottomrule
\end{tabular}
\caption{Per-label performance metrics for the best-performing model on the fine-grained schema (GPT-5-mini, Zero-Shot), sorted by F1-Score. The abbreviations used in the text are provided in parentheses.}
\label{tab:fg_details_full_name}
\end{table*}
\section{Discussion}
\subsection{Main Findings}
Our experimental results provide a quantitative assessment of the capabilities and limitations of modern LLMs for Negative Precedent Treatment Classification. Our primary finding is that model performance is overwhelmingly dictated by the class distribution of the dataset, a classic consequence of class imbalance. This is clearly demonstrated by the best-performing model on the high-level schema, Gemini 2.5 Flash, which achieved its high accuracy (0.7908) by excelling on the most frequent labels. As shown in Table \ref{tab:hl_details}, the LIMITED OR DISTINGUISHED label, constituting a majority of the dataset with 156 instances, was classified with near-perfect precision (0.947) and a strong F1-Score of 0.864. Conversely, the model struggled with less represented labels like NEUTRAL CITATION (10 instances) and CONFLICT NOTED (16 instances), which had poor F1-Scores of just 0.400 and 0.480, respectively.

This challenge is magnified when examining the fine-grained schema, which reveals the limits of applying LLMs to highly specialized, domain-specific taxonomies. The results in Table 7 are stark: seven of the fifteen labels show zero successful predictions, with F1-Scores of 0.000, including legally significant treatments like \textit{Reversed by} and \textit{Questioned by}. This widespread failure suggests that the semantic distinctions provided by commercial citators are often too subtle for models to reliably differentiate from limited data. For a legal practitioner, the practical difference between a precedent being \textit{Criticized by} versus \textit{Questioned by} can be marginal. This poor performance validates our decision to construct the high-level schema, which merges semantically adjacent labels to create a benchmark that is more tractable for current models and more aligned with a realistic legal analysis.

However, even within the more robust high-level schema, classification difficulty varies by label. While the model performed well on INVALIDATED (0.764 F1-Score), its performance on CRITICIZED OR QUESTIONED was weaker (0.651 F1-Score) despite more support (49 instances), suggesting greater ambiguity in the latter. The model's poor performance on NEUTRAL CITATION (0.400 F1-Score) is the most revealing, as it highlights a limitation in our task design. By framing the problem as a choice among predominantly negative labels, the model is biased against selecting the NEUTRAL category. This aligns with known challenges in machine learning where a class is defined by the absence of the primary signal shared by the majority classes, making it a de facto background class that single-stage classifiers often under-select \citep{sillaSurveyHierarchicalClassification2011}.

\subsection{Semantic Overlap and the Nature of Legal Ground Truth}
Our analysis also highlights a fundamental challenge in legal NLP: the thin semantic lines between classification labels. The distinction between a court criticizing versus questioning a precedent, for example, is highly context-dependent and can be subjective. This challenge is amplified by the nature of our source dataset. As \citet*{hellyerEvaluatingShepardsKeyCite2018a} conducted his study not to create a single, definitive ground truth, but to evaluate the inter-accuracy of commercial citator services, his annotations reflect a more flexible standard of what is "acceptable."

Subjectivity is a problem inherent to legal classification tasks, as even legal experts can hold differing views on the correct interpretation of a case, particularly when classifications must be made from semantically nuanced descriptions \citep{kurniawanAggregateNotAggregate2024}. This has profound implications for our evaluation, especially for the fine-grained labels. It means that some classifications marked as "incorrect" by our metrics might still be considered "not wrong" from a legal perspective. An LLM's prediction of \textit{Criticized By} for a case labeled \textit{Disagreed With By} is an error in our benchmark, but it demonstrates a correct grasp of the underlying negative sentiment. Therefore, our results should be interpreted as an exploration of current LLM capabilities and a measure of their alignment with this specific expert-annotated benchmark, rather than a definitive judgment on their legal reasoning. The inherent subjectivity of legal interpretation suggests that there is a ceiling to how high any model's accuracy can be on this task.
\subsection{A Severity Scale for Evaluating Precedent Treatment}
Finally, our analysis reveals that standard accuracy is an insufficient metric for this task, as it treats all misclassifications equally. A model that confuses a case-ending INVALIDATED treatment with a NEUTRAL CITATION makes a far more critical mistake than one that confuses two similar negative labels. To address this, we developed a more nuanced evaluation framework based on a Severity Scale, introduced in Table \ref{tab:severity_scale}, which assigns a score from 1 (NEUTRAL CITATION) to 5 (INVALIDATED). 

This allows us to calculate an Average Severity Error. For our fine-grained evaluation, labels were mapped to their high-level parent category before calculating this error, reinforcing the high-level schema as the core basis for semantic evaluation. Furthermore, to provide a more robust measure of a model's typical performance on this ordinal scale, we also report the Median Severity Error.

The results from this analysis are presented in Table \ref{tab:severity_error_results}. The metric's value is evident: Gemini 2.5 Flash (zero-shot) again proved to be the top-performing model, achieving the lowest Average Severity Error on both the high-level (0.3933) and fine-grained (0.3755) tasks. The colors in the table highlight the Median Severity Error. Critically, the green values indicate a median of 0.0, a significant finding which means that for over half of all predictions, the top-performing models produced the perfectly correct label with zero error. By focusing on both the average magnitude of error and the typical error, our framework provides a more realistic assessment of a model’s fitness for the high-stakes legal domain.

\begin{table}[h!]
    \centering
        \begin{tabular}{@{}lc@{}}
            \toprule
            \textbf{Severity} & \textbf{High-Level Category} \\
            \midrule
            5  & INVALIDATED \\
            4  & CRITICIZED OR QUESTIONED \\
            3  & LIMITED OR DISTINGUISHED \\
            2 & CONFLICT NOTED \\
            1 & NEUTRAL CITATION \\
            \bottomrule
        \end{tabular}
\caption{The severity scale used to evaluate the magnitude of classification errors.}
\label{tab:severity_scale}

\end{table}
\begin{table}[h!]
\centering

\begin{tabular}{@{}ll@{\hspace{0.05em}}cc@{}}
\toprule
\multirow{2}{*}{\textbf{Model}} & \multirow{2}{*}{\textbf{Prompt Type}} & \multicolumn{2}{c}{\textbf{Avg. Error}} \\
\cmidrule(l){3-4}
 & & \textbf{HL} & \textbf{FG} \\
\midrule
\multirow{2}{*}{Gemini 2.5 Flash} & Zero-Shot & \textcolor{green!50!black}{\textbf{0.3933}} & \textcolor{green!50!black}{\textbf{0.3755}} \\
 & Few-Shot & \textcolor{green!50!black}{0.4310} & \textcolor{green!50!black}{0.4686} \\
\midrule
\multirow{2}{*}{Gemini 2.5 Pro} & Zero-Shot & \textcolor{green!50!black}{0.5816} & \textcolor{green!50!black}{0.4192} \\
 & Few-Shot & \textcolor{green!50!black}{0.5069} & \textcolor{green!50!black}{0.4654} \\
\midrule
\multirow{2}{*}{GPT-5-mini} & Zero-Shot & \textcolor{green!50!black}{0.5279} & \textcolor{green!50!black}{0.4260} \\
 & Few-Shot & \textcolor{green!50!black}{0.5945} & \textcolor{red}{0.6544} \\
\midrule
\multirow{2}{*}{Qwen3:30B} & Zero-Shot & \textcolor{green!50!black}{0.5356} & \textcolor{red}{0.5732} \\
 & Few-Shot & \textcolor{green!50!black}{0.7051} & \textcolor{red}{0.5300} \\
\bottomrule
\end{tabular}
\caption{Model performance by Average Severity Error. Values are colored based on their Median Severity Error: \textcolor{green!50!black}{green} for a median of 0.0 (typically perfect) and \textcolor{red}{red} for a median > 0.0 (typically imperfect). Lower scores are better.}
\label{tab:severity_error_results}

\end{table}

\subsection{Qualitative Error Analysis}
To understand the nuances of the model's performance beyond quantitative scores, we conducted a qualitative error analysis. Our prompting strategy required the LLM to provide its reasoning and a verbatim excerpt supporting each prediction, enabling a transparent review of its decision-making process (see Appendix \ref{sec:prompts}). This analysis revealed several key patterns, which we present below.
\subsubsection{Justifiable Disagreement Due to Label Ambiguity}
A primary example of such justifiable disagreements occurred in the treatment of \textit{\citet{Matthews1984}}, where our model predicted CRITICIZED OR QUESTIONED while the ground truth was \textit{Not Followed} for fine-grained and LIMITED OR DISTINGUISHED for high-level labels. The model's prediction was highly defensible, as it correctly identified the citing court’s description of the Matthews rule as leading to “\textit{unintended and inequitable results}”\textit{\citep{Hatfield1990}}. However, the ground truth label is also valid, as it reflects the court's ultimate action of refusing to follow the precedent. This highlights a classic scenario where explicit criticism serves as the justification for limiting a precedent's application, making both labels defensible and classifying the discrepancy as a valid interpretive difference rather than a model error.
\subsubsection{Misattributing the Target of Judicial Action}
A subtle error pattern involves the model misattributing the target of a judicial action. For instance, in the treatment of \textit{\citet{Kail1984}}, the model predicted LIMITED OR DISTINGUISHED against a ground truth of NEUTRAL CITATION, reasoning that the court was "\textit{correcting a party's overbroad interpretation and thereby narrowing its perceived applicability.}" While the model correctly identified this narrowing function, it failed to recognize that the target of the correction was the claimant's argument, not the Kail precedent itself. This misattribution occurred despite our explicit instruction to focus solely on the treatment of the seed case (see Appendix \ref{sec:prompts}). The court actually treats Kail neutrally, highlighting a key challenge for the model: distinguishing between the rhetorical use of a citation and the direct treatment of its legal holding.

\section{Conclusion}
In this paper, we conducted a comprehensive evaluation of modern LLMs for the task of Negative Precedent Treatment Classification. Our findings show that leading proprietary models have significant potential for automating this crucial aspect of legal research. Gemini 2.5 Flash was the top performer on the high-level schema with 79.1\% accuracy, while GPT-5-mini performed best on the more challenging fine-grained schema at 67.7\% accuracy. Despite this promise, overall performance remains constrained by the dataset's class imbalance and the complexity of legal reasoning.

We identify two primary challenges: a scarcity of context-rich legal data and the task's inherent semantic complexity, which pushes the limits of current models. Critically, we find standard accuracy metrics insufficient for this domain. Our proposed Average Severity Error provides a more robust evaluation of model reliability, essential for trustworthy legal AI. We establish a vital baseline and provide a new dataset to the community to advance research in this area.

The dataset used in this work is publicly available on the Hugging Face Hub at \url{https://huggingface.co/datasets/mmikaildemir/negative_treatment}.
\section*{Limitations and Future Work}
While comprehensive, our study is limited in scope. We evaluated a select group of models, and future work should provide a more complete picture by benchmarking a wider variety of systems. Our focus on proprietary models from Google and OpenAI was guided by the availability of research credits, a practical constraint common in academic research. 

The generalizability of our findings also needs to be tested, as our dataset was derived from a single legal jurisdiction. Methodologically, our focus on basic prompting strategies highlights an opportunity for future research to explore more advanced techniques. Investigating fine-tuning and sophisticated context-handling methods, such as retrieval-augmented generation (RAG), will be key to improving the efficiency and accuracy of models on complex legal reasoning tasks.

A critical decision in our setup was to provide the LLM with the full, cleaned text of the citing case as its context, rather than a pre-selected snippet. This choice was made to create the most realistic and challenging testbed. Legal reasoning for a citation's treatment is often not localized to a single sentence; it can be distributed across paragraphs, depend on surrounding arguments, or even be implied by the structure of the legal analysis \cite{panagisGivingEveryCase2017}.

Furthermore, crucial context can be located in non-standard parts of the text, such as footnotes. During our analysis, we identified instances where the determinative information for a classification was present only in a footnote attached to the sentence containing the citation. By providing the full text, we force the model to engage in a more holistic form of document comprehension, requiring it to locate and synthesize the most relevant evidence from a large and complex input. Understanding how models perform on such a complex task requires more than quantitative metrics alone; while our primary focus has been on benchmarking, we provide an initial qualitative analysis to pave the way for the more thorough investigation needed to use these models confidently.
\hypertarget{sec:limitations}{} 
\section*{Acknowledgements}
We are deeply grateful to Paul Hellyer for his essential contributions to this research. This work would not have been possible without his willingness to share his original annotated dataset. Furthermore, we thank him for his invaluable domain expertise and for providing insightful feedback during the review of this paper.
\bibliography{latex/custom}

\clearpage
\appendix

\section{Prompts}
\label{sec:prompts}
\begin{figure}[hbt!]
    \includegraphics[scale=0.13]{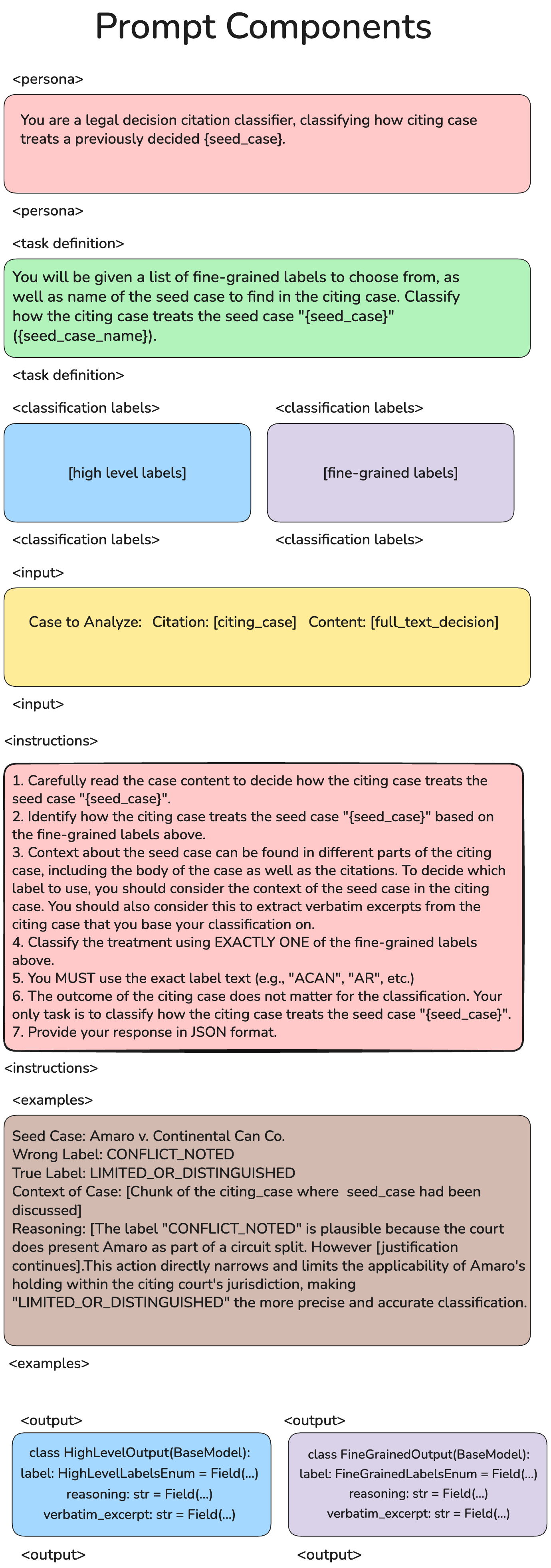}    \caption{An illustration of the prompt components used for classifying legal citation treatments. The architecture includes sections for persona, task definition, classification labels, input, optional examples for few-shot learning, and output specifications for both high-level and fine-grained labeling schemas.
}
    \label{fig:prompts}
\end{figure}

\section{Dataset Example}
\label{sec:B}
\begin{figure*}[h!]
    \centering
    \includegraphics[width=\textwidth]{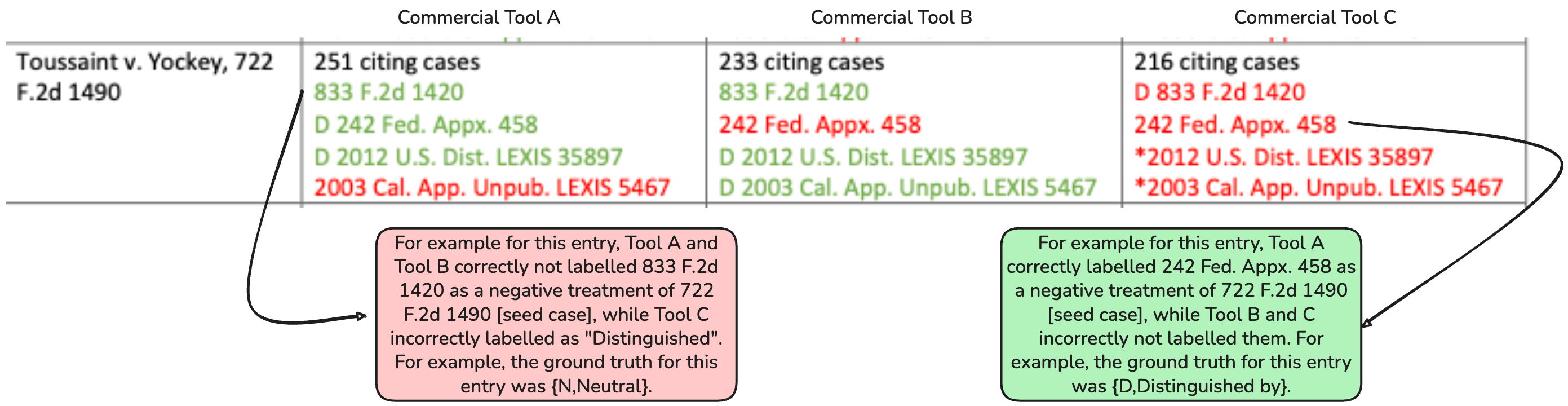}     
    \caption{A snippet of the dataset that \cite{hellyerEvaluatingShepardsKeyCite2018a} provided, with explanations about ground truth logic.
}
    \label{fig:dataset_example}
\end{figure*}
\begin{figure*}
    \centering
    \includegraphics[width=\textwidth]{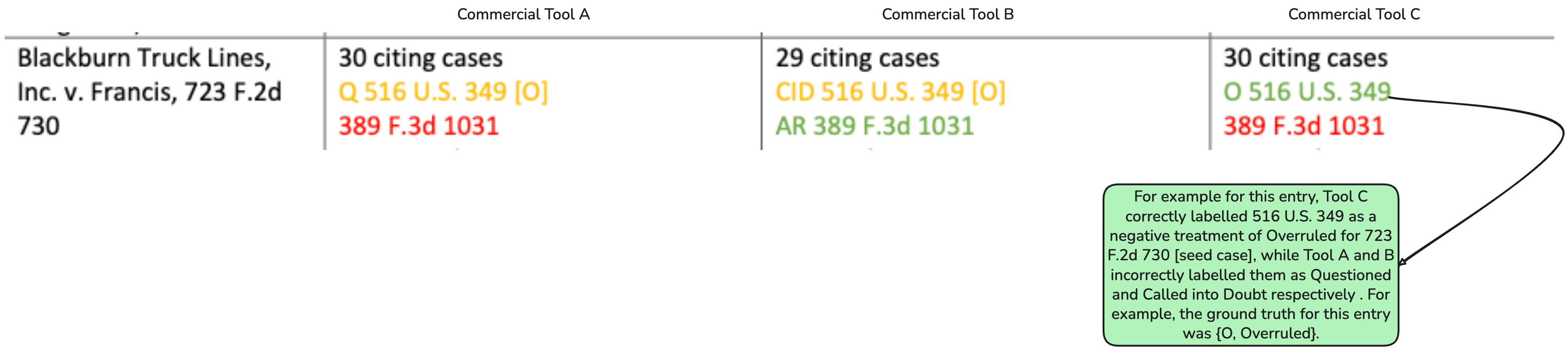}     
    \caption{A snippet of the dataset that \cite{hellyerEvaluatingShepardsKeyCite2018a} provided, where corrected label provided in the brackets
}
    \label{fig:dataset_example_bracket}
\end{figure*}

\end{document}